\documentclass[11pt,a4paper]{article}
\usepackage[hyperref]{acl2018}
\usepackage{times}
\usepackage{latexsym}

\usepackage{url}

\usepackage{amsmath}
\usepackage{amssymb}

\usepackage[linesnumbered,vlined,ruled,norelsize]{algorithm2e}

\aclfinalcopy 
\def\aclpaperid{57} 

\title{Automated essay scoring with string kernels and word embeddings}

\author{M\u{a}d\u{a}lina Cozma \and Andrei M. Butnaru \and {Radu Tudor} Ionescu\\
  University of Bucharest\\
  Department of Computer Science\\
  14 Academiei, Bucharest, Romania\\
  {\tt butnaruandreimadalin@gmail.com}\\
  {\tt raducu.ionescu@gmail.com}
}

\date{}

\begin{document}
\maketitle
\begin{abstract}
In this work, we present an approach based on combining string kernels and word embeddings for automatic essay scoring. String kernels capture the similarity among strings based on counting common character n-grams, which are a low-level yet powerful type of feature, demonstrating state-of-the-art results in various text classification tasks such as Arabic dialect identification or native language identification. To our best knowledge, we are the first to apply string kernels to automatically score essays. We are also the first to combine them with a high-level semantic feature representation, namely the bag-of-super-word-embeddings. We report the best performance on the Automated Student Assessment Prize data set, in both in-domain and cross-domain settings, surpassing recent state-of-the-art deep learning approaches.
\end{abstract}

\setlength{\abovedisplayskip}{3pt}
\setlength{\belowdisplayskip}{3pt}

\section{Introduction}

Automatic essay scoring (AES) is the task of assigning grades to essays written in an educational setting, using a computer-based system with natural language processing capabilities. The aim of designing such systems is to reduce the involvement of human graders as far as possible. AES is a challenging task as it relies on grammar as well as semantics, pragmatics and discourse \cite{Song-ACL-2017}. Although traditional AES methods typically rely on handcrafted features \cite{Larkey-SIGIR-1998,Foltz-EdMedia-1999,Attali-JTLA-2006,Dikli-JTLA-2006,Wang-CITTE-2008,Chen-EMNLP-2013,Somasumdaran-COLING-2014,Yannakoudakis-ACL-2014,Phandi-EMNLP-2015}, recent results indicate that state-of-the-art deep learning methods reach better performance \cite{Alikaniotis-ACL-2016,Dong-EMNLP-2016,Taghipour-EMNLP-2016,Dong-CONLL-2017,Song-ACL-2017,Tay-ACL-2018}, perhaps because these methods are able to capture subtle and complex information that is relevant to the task \cite{Dong-EMNLP-2016}.

In this paper, we propose to combine string kernels (low-level character n-gram features) and word embeddings (high-level semantic features) to obtain state-of-the-art AES results. Since recent methods based on string kernels have demonstrated remarkable performance in various text classification tasks ranging from authorship identification \cite{PopescuG12} and sentiment analysis \cite{franco-EACL-2017,marius-KES-2017} to native language identification \cite{popescu-ionescu:2013:BEA8,ionescu-popescu-cahill-EMNLP-2014,Radu-ICONIP-2015,ionescu-popescu-cahill-COLI-2016,Ionescu-BEA-2017} and dialect identification \cite{Radu-Marius-ADI-2016,Radu-Andrei-ADI-2017}, we believe that string kernels can reach equally good results in AES. To the best of our knowledge, string kernels have never been used for this task. As string kernels are a simple approach that relies solely on character n-grams as features, it is fairly obvious that such an approach will not to cover several aspects (e.g.: semantics, discourse) required for the AES task. To solve this problem, we propose to combine string kernels with a recent approach based on word embeddings, namely the bag-of-super-word-embeddings (BOSWE) \cite{Ionescu-KES-2017}. To our knowledge, this is the first successful attempt to combine string kernels and word embeddings. We evaluate our approach on the Automated Student Assessment Prize data set, in both in-domain and cross-domain settings. The empirical results indicate that our approach yields a better performance than state-of-the-art approaches \cite{Phandi-EMNLP-2015,Dong-EMNLP-2016,Dong-CONLL-2017,Tay-ACL-2018}.

\vspace*{-0.1cm} 
\section{Method}
\label{sec_String_Kernels}
\vspace*{-0.05cm} 

\noindent
{\bf String kernels.}
Kernel functions \cite{taylor-Cristianini-cup-2004} capture the intuitive notion of similarity between objects in a specific domain. For example, in text mining, string kernels can be used to measure the pairwise similarity between text samples, simply based on character n-grams. Various string kernel functions have been proposed to date \cite{LodhiSSCW02,taylor-Cristianini-cup-2004,ionescu-popescu-cahill-EMNLP-2014}. One of the most recent string kernels is the \emph{histogram intersection string kernel} (HISK) \cite{ionescu-popescu-cahill-EMNLP-2014}. For two strings over an alphabet $\Sigma$, $x,y \in \Sigma^*$, the intersection string kernel is formally defined as follows:
\begin{equation}
\begin{split}
k^{\cap}(x,y)=\sum\limits_{v \in \Sigma^n} \min \lbrace \mbox{num}_v(x), \mbox{num}_v(y) \rbrace ,
\end{split}
\end{equation}
where $\mbox{num}_v(x)$ is the number of occurrences of n-gram $v$ as a substring in $x$, and $n$ is the length of $v$. In our AES experiments, we use the intersection string kernel based on a range of character n-grams. We approach AES as a regression task, and employ $\nu$-Support Vector Regression ($\nu$-SVR) \cite{Suykens-NPL-1999,taylor-Cristianini-cup-2004} for training.

\noindent
{\bf Bag-of-super-word-embeddings.}
Word embeddings are long known in the NLP community \cite{Bengio-JMLR-2003,Collobert-ICML-2008}, but they have recently become more popular due to the \emph{word2vec} \cite{Mikolov-NIPS-2013} framework that enables the building of efficient vector representations from words. On top of the word embeddings, \newcite{Ionescu-KES-2017} developed an approach termed \emph{bag-of-super-word-embeddings} (BOSWE) by adapting an efficient computer vision technique, the bag-of-visual-words model \cite{Csurka-2004}, for natural language processing tasks. The adaptation consists of replacing the image descriptors \cite{Lowe-SIFT-2004} useful for recognizing object patterns in images with word embeddings \cite{Mikolov-NIPS-2013} useful for recognizing semantic patterns in text documents. 

The BOSWE representation is computed as follows. First, each word in the collection of training documents is represented as word vector using a pre-trained word embeddings model. Based on the fact that word embeddings carry semantic information by projecting semantically related words in the same region of the embedding space, the next step is to cluster the word vectors in order to obtain relevant semantic clusters of words. As in the standard bag-of-visual-words model, the clustering is done by k-means \cite{Leung-2001}, and the formed centroids are stored in a randomized forest of k-d trees \cite{Philbin-2007} to reduce search cost. The centroid of each cluster is interpreted as a \emph{super word embedding} or \emph{super word vector} that embodies all the semantically related word vectors in a small region of the embedding space. Every embedded word in the collection of documents is then assigned to the nearest cluster centroid (the nearest super word vector). Put together, the super word vectors generate a vocabulary (codebook) that can further be used to describe each document as a \emph{bag-of-super-word-embeddings}. To obtain the BOSWE represenation for a document, we just have to compute the occurrence count of each super word embedding in the respective document. After building the representation, we employ a kernel method to train the BOSWE model for our specific task. To be consistent with the string kernel approach, we choose the histogram intersection kernel and the same regression method, namely $\nu$-SVR.

\noindent
{\bf Model fusion.}
In the primal form, a linear classifier takes as input a feature matrix $X$ of $r$ samples (rows) with $m$ features (columns) and optimizes a set of weights in order to reproduce the $r$ training labels. In the dual form, the linear classifier takes as input a kernel matrix $K$ of $r \times r$ components, where each component $k_{ij}$ is the similarity between examples $x_i$ and $x_j$. Kernel methods work by embedding the data in a Hilbert space and by searching for linear relations in that space, using a learning algorithm. The embedding can be performed either $(i)$ implicitly, by directly specifying the similarity function between each pair of samples, or $(ii)$ explicitly, by first giving the embedding map $\phi$ and by computing the inner product between each pair of samples embedded in the Hilbert space. For the linear kernel, the associated embedding map is $\phi(x) = x$ and options $(i)$ or $(ii)$ are equivalent, i.e. the similarity function is the inner product. Hence, the linear kernel matrix $K$ can be obtained as $K = X \cdot X'$, where $X'$ is the transpose of $X$. For other kernels, e.g. the histogram intersection kernel, it is not possible to explicitly define the embedding map  \cite{taylor-Cristianini-cup-2004}, and the only solution is to adopt option $(i)$ and compute the corresponding kernel matrix directly. Therefore, we combine HISK and BOSWE in the dual (kernel) form, by simply summing up the two corresponding kernel matrices. However, summing up kernel matrices is equivalent to feature vector concatenation in the primal Hilbert space. To better explain this statement, let us suppose that we can define the embedding map of the histogram intersection kernel and, consequently, we can obtain the corresponding feature matrix of HISK with $r \times m_1$ components denoted by $X_1$ and the corresponding feature matrix of BOSWE with $r \times m_2$ components denoted by $X_2$. We can now combine HISK and BOSWE in two ways. One way is to compute the corresponding kernel matrices $K_1$ = $X_1 \cdot X_1'$ and $K_2 = X_2 \cdot X_2'$, and to sum the matrices into a single kernel matrix $K_+ = K_1 + K_2$. The other way is to first concatenate the feature matrices into a single feature matrix $X_+ = [X_1 X_2]$ of $r \times (m_1 + m_2)$ components, and to compute the final kernel matrix using the inner product, i.e. $K_+ = X_+ \cdot X_+'$. Either way, the two approaches, HISK and BOSWE, are fused before the learning stage. As a consequence of kernel summation, the search space of linear patterns grows, which should help the kernel classifier, in our case $\nu$-SVR, to find a better regression function.

\vspace*{-0.1cm} 
\section{Experiments}
\label{sec_Polarity_Experiments}

\begin{table}[!t]
\setlength\tabcolsep{5.5pt}
\begin{center}
\begin{tabular}{ccc}
\hline
Prompt 				& Number of Essays		&	Score Range	\\
\hline
\hline
\vspace{-0.9em}\\
1							& 1783						& 2-12\\
2							& 1800 						& 1-6\\											
3							& 1726						& 0-3\\													
4							& 1726						& 0-3\\																							
5							& 1772						& 0-4\\
6							& 1805						& 0-4\\
7							& 1569						& 0-30\\
8							& 723						& 0-60\\
\hline
\end{tabular}
\end{center}
\vspace*{-0.2cm}
\caption{The number of essays and the score ranges for the 8 different prompts in the Automated Student Assessment Prize (ASAP) data set.}
\label{tab_asap}
\vspace*{-0.3cm}
\end{table}

\begin{table*}[!t]
\setlength\tabcolsep{3.5pt}
\begin{center}
\begin{tabular}{lccccccccc}
\hline
Method 											& 1								&	2								& 3								& 4 						
					& 5 								& 6								& 7 								& 8 								& Overall\\
\hline
\hline
\vspace{-0.9em}\\
Human												& $0.721$ 					& $0.814$ 					& $0.769$ 					& $0.851$
					& $0.753$ 					& $0.776$ 					& $0.721$					& $0.629$					& $0.754$\\
\hline
\vspace{-0.9em}\\
\cite{Phandi-EMNLP-2015}				& $0.761$ 					& $0.606$ 					& $0.621$ 					& $0.742$
					& $0.784$ 					& $0.775$ 					& $0.730$					& $0.617$					& $0.705$\\
															
\cite{Dong-EMNLP-2016}				& - 								& - 								& - 								& -
					& - 								& - 								& -								& -								& $0.734$\\
															
\cite{Dong-CONLL-2017}				& $0.822$ 					& $0.682$ 					& $0.672$ 					& $0.814$
					& $0.803$ 					& $0.811$ 					& $0.801$					& $0.705$ 					& $0.764$\\
																									
\cite{Tay-ACL-2018}						& $0.832$ 					& $0.684$ 					& $\mathbf{0.695}$ 	& $0.788$
					& $0.815$ 					& $0.810$ 					& $0.800$					& $0.697$					& $0.764$\\
\hline
\vspace{-0.9em}\\
HISK and $\nu$-SVR						& $0.836$ 					& $0.724$ 					& $0.677$ 					& $0.821$
					& $0.830$ 					& $0.828$ 					& $0.801$					& $0.726$					& $0.780$\\
																					
BOSWE and $\nu$-SVR					& $0.788$ 					& $0.689$ 					& $0.667$ 					& $0.809$
					& $0.824$ 					& $0.824$ 					& $0.766$					& $0.679$					& $0.756$\\
																					
HISK+BOSWE and $\nu$-SVR			& $\mathbf{0.845}$ 	& $\mathbf{0.729}$ 	& $0.684$ 					& $\mathbf{0.829}$
					& $\mathbf{0.833}$	& $\mathbf{0.830}$	& $\mathbf{0.804}$	& $\mathbf{0.729}$	& $\mathbf{0.785}$\\
\hline
\end{tabular}
\end{center}
\vspace*{-0.2cm}
\caption{In-domain automatic essay scoring results of our approach versus several state-of-the-art methods \cite{Phandi-EMNLP-2015,Dong-EMNLP-2016,Dong-CONLL-2017,Tay-ACL-2018}. Results are reported in terms of the quadratic weighted kappa (QWK) measure, using 5-fold cross-validation. The best QWK score (among the machine learning systems) for each prompt is highlighted in bold.}
\label{tab_AES_in}
\vspace*{-0.3cm}
\end{table*}

\noindent
{\bf Data set.}
To evaluate our approach, we use the Automated Student Assessment Prize (ASAP) \footnote{\scriptsize{\url{https://www.kaggle.com/c/asap-aes/data}}} data set from Kaggle. The ASAP data set contains 8 prompts of different genres. The number of essays per prompt along with the score ranges are presented in Table~\ref{tab_asap}. Since the official test data of the ASAP competition is not released to the public, we, as well as others before us \cite{Phandi-EMNLP-2015,Dong-EMNLP-2016,Dong-CONLL-2017,Tay-ACL-2018}, use only the training data in our experiments.

\noindent
{\bf Evaluation procedure.}
As \newcite{Dong-EMNLP-2016}, we scaled the essay scores into the range 0-1. We closely followed the same settings for data preparation as \cite{Phandi-EMNLP-2015,Dong-EMNLP-2016}. For the in-domain experiments, we use 5-fold cross-validation. The 5-fold cross-validation procedure is repeated for 10 times and the results were averaged to reduce the accuracy variation introduced by randomly selecting the folds. We note that the standard deviation in all cases in below $0.2\%$.

For the cross-domain experiments, we use the same source$\rightarrow$target domain pairs as \cite{Phandi-EMNLP-2015,Dong-EMNLP-2016}, namely, 1$\rightarrow$2, 3$\rightarrow$4, 5$\rightarrow$6 and 7$\rightarrow$8. All essays in the source domain are used as training data. Target domain samples are randomly divided into 5 folds, where one fold is used as test data, and the other 4 folds are collected together to sub-sample target domain train data. The sub-sample sizes are $n_t = \{ 10, 25, 50, 100 \}$. The sub-sampling is repeated for 5 times as in \cite{Phandi-EMNLP-2015,Dong-EMNLP-2016} to reduce bias. As our approach performs very well in the cross-domain setting, we also present experiments \emph{without} sub-sampling data from the target domain, i.e. when the sub-sample size is $n_t=0$. As evaluation metric, we use the quadratic weighted kappa (QWK).

\noindent
{\bf Baselines.}
We compare our approach with state-of-the-art methods based on handcrafted features \cite{Phandi-EMNLP-2015}, as well as deep features \cite{Dong-EMNLP-2016,Dong-CONLL-2017,Tay-ACL-2018}. We note that results for the cross-domain setting are reported only in some of these recent works \cite{Phandi-EMNLP-2015,Dong-EMNLP-2016}.

\noindent
{\bf Implementation choices.}
For the string kernels approach, we used the histogram intersection string kernel (HISK) based on the blended range of character n-grams from 1 to 15. To compute the intersection string kernel, we used the open-source code provided by \newcite{ionescu-popescu-cahill-EMNLP-2014}. For the BOSWE approach, we used the pre-trained word embeddings computed by the \emph{word2vec} toolkit \cite{Mikolov-NIPS-2013} on the Google News data set using the Skip-gram model, which produces $300$-dimensional vectors for $3$ million words and phrases. We used functions from the VLFeat library \cite{vedaldi-vlfeat-2008} for the other steps involved in the BOSWE approach, such as the k-means clustering and the randomized forest of k-d trees. We set the number of clusters (dimension of the vocabulary) to $k = 500$. After computing the BOSWE representation, we apply the $L_1$-normalized intersection kernel. We combine HISK and BOSWE in the dual form by summing up the two corresponding matrices. For the learning phase, we employ the dual implementation of $\nu$-SVR available in LibSVM \cite{LibSVM-2011}. We set its regularization parameter to $c=10^3$ and $\nu = 10^{-1}$ in all our experiments.

\begin{table*}[!t]
\setlength\tabcolsep{4.5pt}
\begin{center}
\begin{tabular}{clccccc}
\hline
Source$\rightarrow$Target & Method 														& $n_t=0$					&	$n_t=10$					
																				&	$n_t=25$				& $n_t=50$					& $n_t=100$\\
\hline
\hline
\vspace{-0.9em}\\
1$\rightarrow$2					& \cite{Phandi-EMNLP-2015}							& $0.434$ 					& $0.463$ 				
																				& $0.457$ 					& $0.492$					& $0.510$\\
																				
											& \cite{Dong-EMNLP-2016}								& - 								& $0.546$ 				
																				& $0.569$ 					& $0.563$					& $0.559$\\
\cline{2-7}
\vspace{-0.9em}\\

											& HISK and $\nu$-SVR										& $0.440$ 					& $\mathbf{0.586}$ 				
																				& $\mathbf{0.637}$ 	& $0.652$					& $0.657$\\

											& BOSWE and $\nu$-SVR								& $0.398$ 					& $0.474$ 				
																				& $0.478$ 					& $0.492$					& $0.506$\\

											& HISK+BOSWE and $\nu$-SVR						& $\mathbf{0.542}$ 	& $0.584$ 				
																				& $0.632$ 					& $\mathbf{0.657}$	& $\mathbf{0.661}$\\
\hline
\vspace{-0.9em}\\
3$\rightarrow$4					& \cite{Phandi-EMNLP-2015}							& $0.522$ 					& $0.593$ 				
																				& $0.609$ 					& $0.618$					& $0.646$\\
																				
											& \cite{Dong-EMNLP-2016}								& - 								& $0.628$ 				
																				& $0.656$ 					& $0.659$					& $0.662$\\
\cline{2-7}
\vspace{-0.9em}\\

											& HISK and $\nu$-SVR										& $\mathbf{0.703}$ 	& $\mathbf{0.716}$ 				
																				& $0.724$ 					& $0.742$					& $0.751$\\

											& BOSWE and $\nu$-SVR								& $0.615$ 					& $0.640$ 				
																				& $0.716$ 					& $0.728$					& $0.727$\\

											& HISK+BOSWE and $\nu$-SVR						& $0.701$ 					& $0.713$ 				
																				& $\mathbf{0.737}$ 	& $\mathbf{0.754}$	& $\mathbf{0.779}$\\
\hline
\vspace{-0.9em}\\
5$\rightarrow$6					& \cite{Phandi-EMNLP-2015}							& $0.187$ 					& $0.539$ 				
																				& $0.662$ 					& $0.680$					& $0.713$\\
																				
											& \cite{Dong-EMNLP-2016}								& - 								& $0.647$ 				
																				& $0.700$ 					& $0.714$					& $0.750$\\
\cline{2-7}
\vspace{-0.9em}\\

											& HISK and $\nu$-SVR										& $0.715$ 					& $0.726$ 				
																				& $0.754$ 					& $0.757$					& $0.781$\\

											& BOSWE and $\nu$-SVR								& $0.617$ 					& $0.623$ 				
																				& $0.644$ 					& $0.650$					& $0.692$\\

											& HISK+BOSWE and $\nu$-SVR						& $\mathbf{0.728}$ 	& $\mathbf{0.734}$ 				
																				& $\mathbf{0.764}$ 	& $\mathbf{0.771}$	& $\mathbf{0.788}$\\
\hline
\vspace{-0.9em}\\
7$\rightarrow$8					& \cite{Phandi-EMNLP-2015}							& $0.171$ 					& $0.586$ 				
																				& $0.607$ 					& $0.613$					& $0.621$\\
																				
											& \cite{Dong-EMNLP-2016}								& - 								& $0.570$ 				
																				& $0.590$ 					& $0.568$					& $0.587$\\
\cline{2-7}
\vspace{-0.9em}\\

											& HISK and $\nu$-SVR										& $0.486$ 					& $0.604$ 				
																				& $0.617$ 					& $0.626$					& $0.639$\\

											& BOSWE and $\nu$-SVR								& $0.419$ 					& $0.526$ 				
																				& $0.577$ 					& $0.582$					& $0.591$\\

											& HISK+BOSWE and $\nu$-SVR						& $\mathbf{0.522}$ 	& $\mathbf{0.606}$ 				
																				& $\mathbf{0.637}$ 	& $\mathbf{0.638}$	& $\mathbf{0.649}$\\
\hline
\end{tabular}
\end{center}
\vspace*{-0.2cm}
\caption{Corss-domain automatic essay scoring results of our approach versus two state-of-the-art methods \cite{Phandi-EMNLP-2015,Dong-EMNLP-2016}. Results are reported in terms of the quadratic weighted kappa (QWK) measure, using the same evaluation procedure as \cite{Phandi-EMNLP-2015,Dong-EMNLP-2016}. The best QWK scores for each source$\rightarrow$target domain pair are highlighted in bold.}
\label{tab_AES_cross}
\vspace*{-0.3cm}
\end{table*}

\noindent
{\bf In-domain results.}
The results for the in-domain automatic essay scoring task are presented in Table~\ref{tab_AES_in}. In our empirical study, we also include feature ablation results. We report the QWK measure on each prompt as well as the overall average. We first note that the histogram intersection string kernel alone reaches better overall performance ($0.780$) than all previous works \cite{Phandi-EMNLP-2015,Dong-EMNLP-2016,Dong-CONLL-2017,Tay-ACL-2018}. Remarkably, the overall performance of the HISK is also higher than the inter-human agreement ($0.754$). Although the BOSWE model can be regarded as a shallow approach, its overall results are comparable to those of deep learning approaches \cite{Dong-EMNLP-2016,Dong-CONLL-2017,Tay-ACL-2018}. When we combine the two models (HISK and BOSWE), we obtain even better results. Indeed, the combination of string kernels and word embeddings attains the best performance on 7 out of 8 prompts. The average QWK score of HISK and BOSWE ($0.785$) is more than $2\%$ better the average scores of the best-performing state-of-the-art approaches \cite{Dong-CONLL-2017,Tay-ACL-2018}.

\noindent
{\bf Cross-domain results.} The results for the cross-domain automatic essay scoring task are presented in Table~\ref{tab_AES_cross}. For each and every source$\rightarrow$target pair, we report better results than both state-of-the-art methods \cite{Phandi-EMNLP-2015,Dong-EMNLP-2016}. We observe that the difference between our best QWK scores and the other approaches are sometimes much higher in the cross-domain setting than in the in-domain setting. We particularly notice that the difference from \cite{Phandi-EMNLP-2015} when $n_t = 0$ is always higher than $10\%$. Our highest improvement (more than $54\%$, from $0.187$ to $0.728$) over \cite{Phandi-EMNLP-2015} is recorded for the pair 5$\rightarrow$6, when $n_t = 0$. Our score in this case ($0.728$) is even higher than both scores of \newcite{Phandi-EMNLP-2015} and \newcite{Dong-EMNLP-2016} when they use $n_t = 50$. Different from the in-domain setting, we note that the combination of string kernels and word embeddings does not always provide better results than string kernels alone, particularly when the number of target samples ($n_t$) added into the training set is less or equal to 25.

\noindent
{\bf Discussion.} It is worth noting that in a set of preliminary experiments (not included in the paper), we actually considered another approach based on word embeddings. We tried to obtain a document embedding by averaging the word vectors for each document. We computed the average as well as the standard deviation for each component of the word vectors, resulting in a total of $600$ features, since the word vectors are $300$-dimensional. We applied this method in the in-domain setting and we obtained a surprisingly low overall QWK score, around $0.251$. We concluded that this simple approach is not useful, and decided to use BOSWE \cite{Ionescu-KES-2017} instead.

It would have been interesting to present an error analysis based on the discriminant features weighted higher by the $\nu$-SVR method. Unfortunately, this is not possible because our approach works in the dual space and we cannot transform the dual weights into primal weights, as long as the histogram intersection kernel does not have an explicit embedding map associated to it. In future work, however, we aim to replace the histogram intersection kernel with the presence bits kernel, which will enable us to perform an error analysis based on the overused or underused patterns, as described by \newcite{ionescu-popescu-cahill-COLI-2016}.

\vspace*{-0.2cm} 
\section{Conclusion}
\label{sec_Conclusion}
\vspace*{-0.1cm} 

In this paper, we described an approach based on combining string kernels and word embeddings for automatic essay scoring. We compared our approach on the Automated Student Assessment Prize data set, in both in-domain and cross-domain settings, with several state-of-the-art approaches \cite{Phandi-EMNLP-2015,Dong-EMNLP-2016,Dong-CONLL-2017,Tay-ACL-2018}. Overall, the in-domain and the cross-domain comparative studies indicate that string kernels, both alone and in combination with word embeddings, attain the best performance on the automatic essay scoring task. Using a shallow approach, we report better results compared to recent deep learning approaches \cite{Dong-EMNLP-2016,Dong-CONLL-2017,Tay-ACL-2018}.

\section*{Acknowledgments}
We thank the reviewers for their useful comments. The work of Radu Tudor Ionescu was partially supported through project grant PN-III-P1-1.1-PD-2016-0787.

\bibliography{references}
\bibliographystyle{acl_natbib}

\end{document}